# DEMONSTRATING THE DUAL-CHANNEL FEATURE DECOUPLING CHARACTERISTIC COMMON TO FRONT-DOOR INTERVENTION CAUSAL INFERENCE METHODS IN WHOLE SLIDE IMAGE CLASSIFICATION

*Zhirui Zhang[1#] Tianhang Nan[1#], Yong Ding[1], Zhuolun Song[1], Dayu Hu[1*], Xiaoyu Cui[1*]*

*Authors masked with # have the same contribution.*

1. College of Medicine and Biological Information Engineering, Northeastern University, China

## ABSTRACT

Causal inference using front door intervention and multi-instance learning (MIL) has advanced the analysis of Whole Slide Images (WSI) in digital pathology. These methods adjust feature distributions of subtle evidence sub-images to correctly associate them with WSI-level diagnoses. We propose and prove 2 hypotheses for evaluating such methods: 1) Causal inference MIL introduces an independent classification channel that effectively completes WSI classification; 2) Greater difference between features extracted by the new and baseline channels increases effectiveness in eliminating false correlations. This hypothesis describes the core of causal inference MILs: overlaying parallel, independent channels to eliminate false associations between WSI-level diagnostic and non-diagnostic evidence sub-images by increasing deep feature diversity. Based on these hypotheses, we evaluated several causal inference MILs on breast cancer and non-small cell lung cancer datasets. This hypothesis provides a new theoretical perspective for applying causal inference to WSI analysis.

## 1. INTRODUCTION

As the “gold standard” for disease diagnosis, the precision of pathology profoundly influences clinical decision-making [1-3]. The maturation of whole-slide imaging (WSI) technology has propelled histopathological analysis into the digital era, enabling computational analysis of gigapixel-level images. Multiple instance learning (MIL), as a canonical paradigm of weakly supervised learning [4, 5], addresses these challenges by decomposing WSIs into thousands of image patches (instances) and training classification models using only WSI-level labels [6-8]. This approach assumes that at least one critical instance (e.g., a cancerous region) is relevant to the global label and aggregates instance features via attention mechanisms to achieve global predictions[9]. Although MIL significantly reduces annotation requirements, it remains a correlation-based statistical learning method with the following limitations: 1. Traditional aggregators treat all instance features equally, failing to distinguish causal features (e.g., cellular atypia) from confounding features (e.g., staining intensity). 2. When training data exhibit selection bias (e.g., a hospital using different staining protocols for slides of different diseases), models may erroneously solidify non-causal associations (e.g., staining style-diagnosis outcome) as decision-making criteria[10].

Recent advances in front-door intervention-based causal inference have provided a novel pathway for eliminating unobserved confounders in MIL. According to Pearl’s causal hierarchy theory [11], the front-door criterion establishes a causal chain $X \rightarrow M \rightarrow Y$ via a mediator variable, making it particularly suitable for scenarios involving unmeasurable confounders (e.g., institutional staining protocols) [12]. In MIL tasks, this framework treats image patches as input variables $X$. Local histopathological features ($X$) influence diagnostic outcomes ($Y$) through deep learning representations ($M$), while confounding factors such as scanner variability ($Z$), constrained by the feature extractor and aggregator) can only indirectly affect predictions via the $X \rightarrow Y$ pathway (Figure 1.a). By applying the front-door adjustment [11-13], the model can block the confounding path $Z \rightarrow X \rightarrow M \rightarrow Y$ without directly observing $Z$ (Figure 1.b & c).

Although front-door intervention-based causal inference methods provide mathematical guarantees for eliminating unobserved confounders at the theoretical level[14], their clinical application faces significant challenges: pathologists remain skeptical about the interpretability of black-box causal modeling. Existing studies predominantly validate method effectiveness through classification performance metrics (e.g., accuracy, AUC) [15, 16]. However, due to their failure to elucidate the correspondence between causal intervention mechanisms and histopathological evidence, as well as the lack of explainable causal relationships in their results, these studies suffer from insufficient clinical trustworthiness.

By deconstructing the front-door adjustment formula and the forward propagation paths of network parameters, we found that mainstream approaches (e.g., feature re-weighting based on the do-operator[17], stratified sampling of mediator variables[18]) essentially construct a dual-pathway classification architecture: the baseline pathway learns correlation patterns in the original feature space, while the causal pathway extracts deconfounded causal patterns through front-door intervention (Figure 1.d). We establish a universal theoretical framework for front-door intervened MILs grounded in Structural Causal Models (SCM) and Information Theory. We rigorously derive and validate two core principles:

1. The causal pathway can operate independently of the baseline pathway while maintaining diagnostic efficacy. Computing the interventional distribution $P(Y|do(X))$ necessitates a marginalization over a global prior *P(X')*. We prove that this mathematical operation strictly requires the model to bifurcate into a dual-pathway architecture: a baseline channel capturing the biased observational distribution, and an independent causal channel executing the interventional expectation.
2. The degree of feature divergence $\Delta$ between the dual pathways determines the deconfounding effect. Defining the inner product of feature vectors as the inter-channel feature divergence, the causal pathway preferentially attends to instances exhibiting greater WSI-level feature discrepancies compared to the baseline pathway (Figure 1.e). This represents an information-complementary decoupling that mitigates spurious correlations arising from the baseline pathway's overreliance on input dependencies. We demonstrate that the geometric feature divergence $\Delta$ between the dual pathways is not a mere empirical similarity metric, but a strict variational upper bound for confounder isolation. Let $Q$ and $K$ denote the representations extracted by the baseline and causal channels, respectively. According to the Data Processing Inequality (DPI), suppressing the mutual information $I(K;Q)$ stringently limits the entanglement between the causal feature $K$ and the unobserved confounder $Z$. Consequently, we establish the following optimization equivalence:

$$\max \Delta(Q,K) \Leftrightarrow \min I(K;Q) \Rightarrow \min \text{UpperBound}\big(I(K;Z)\big)$$

This formulation quantifies the confounding-blocking effect: maximizing $\Delta$ mathematically forces the causal channel to project features orthogonally to the $Z$-dependent subspace, thereby fulfilling the unconfoundedness prerequisite of the front-door criterion.

A larger $\Delta$ indicates that $M$ has a more significant effect from $Z$ in the baseline pathway, while the causal pathway successfully eliminates the confounding interference. On Camelyon16 (dataset of breast cancer sentinel lymph node metastases with manual pixelwise annotations), these instances exhibited substantial overlap with ground truth annotations, demonstrating that the causal pathway establishes genuine associations between model predictions and diagnostically evidential patches.

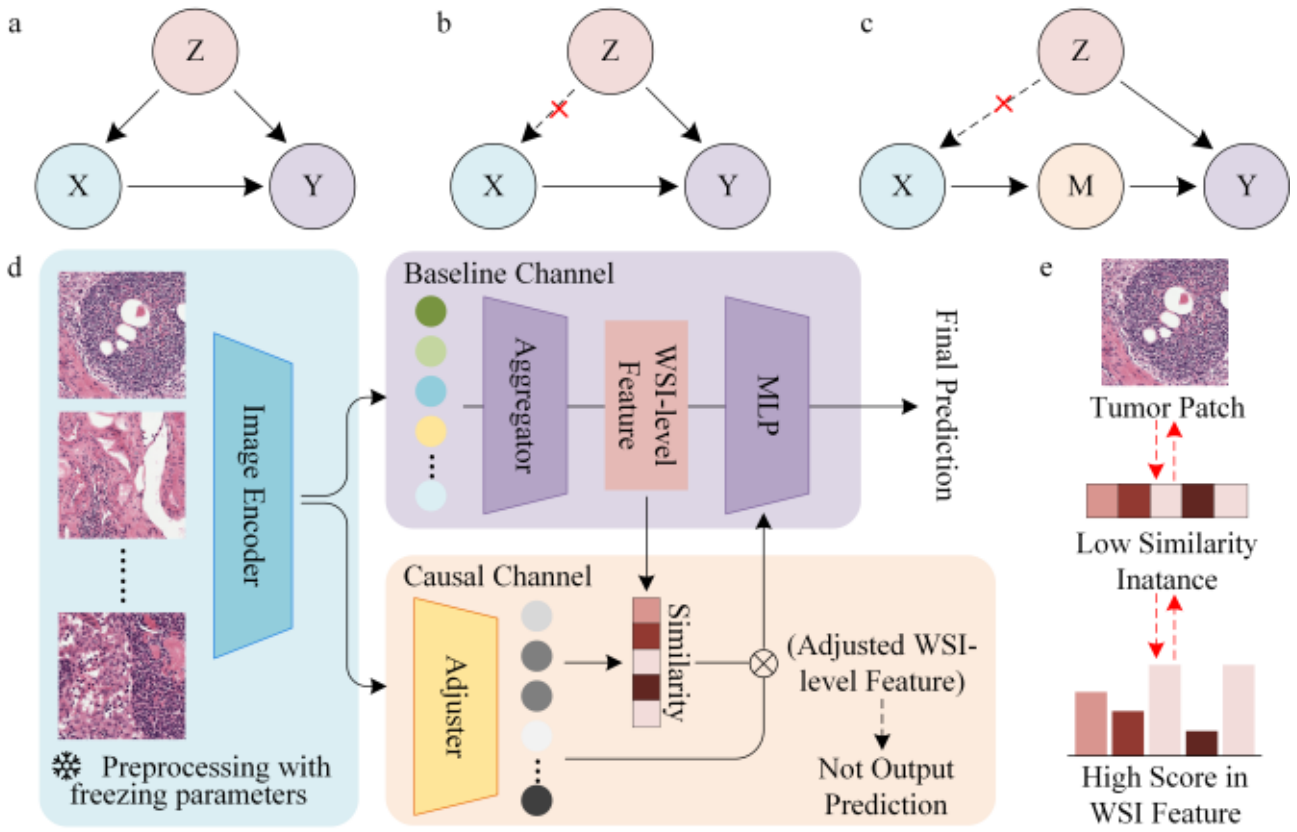

**Figure 1.a-c.** Causal diagrams for no intervention, back-door intervention and front-door intervention. **d.** Workflow of front-door intervention. **e.** Patches relevant to diagnosis exhibit substantial dissimilarity from the WSI-level features in the baseline channel.

These two principles elucidate the underlying mechanism of front-door-intervened causal MIL: by constructing parallel yet orthogonal feature representation spaces, feature diversity suppresses spurious correlations. Unlike conventional methods that solely optimize feature aggregation mechanisms[6, 7], this structural innovation shifts the model's focus from "how to fuse" to "how to decouple", better aligning with the clinical reasoning logic of "critical evidence identification" in pathological diagnosis.

Through systematic validation of the aforementioned dual-channel hypotheses, this study elucidates the interpretability formation mechanism in front-door-intervened causal MIL. Thus, we develop a rigorous and broadly applicable evaluation framework for establishing explanatory rationality in practical pathological image analysis—an aspect that previously lacked standardized assessment. Within this framework, we conducted a unified evaluation of existing front-door-intervened causal MIL methods, demonstrating their effectiveness and interpretability from this novel perspective. The findings reveal both the common limitations and potential improvement directions for front-door intervention methods at the interpretability level, thereby providing new empirical foundations for the emerging field of causality-aware interpretable WSI-assisted diagnosis.

## 2. METHOD

2.1 Backdoor Path in Whole Slice Image Diagnosis Tasks.

In a standard Whole Slide Image (WSI) classification task, let the macroscopic variable $X$ denote the complete pathological image and $Y$ denote the definitive diagnostic outcome (e.g., benign or malignant). In an ideal, unbiased environment devoid of any data acquisition artifacts, the diagnostic prediction relies exclusively on genuine morphological evidence present in the image. This establishes a direct, unconfounded causal pathway from the WSI to the diagnosis: $X \rightarrow Y$.

Under this assumption, the objective of the predictive model is to fit a deterministic function $f(\cdot)$ that maps the WSI directly to the predicted diagnosis $Y$:

$$Y = f(X) \quad (1)$$

In this ideal state, the model's predictive distribution $P(Y|X)$ perfectly aligns with the true causal mechanism, capturing only the pathological variations without any spurious interference.

However, such ideal assumptions may not always hold in real-world observational data. To mathematically formalize the realistic data generating mechanism without introducing ambiguous exogenous noise variables, this task can be described as a causal graph based on causal theory and the Structural Causal Model (SCM) framework as shown in Figure 1.a.

Let $Z$ denote a shared, unobserved variable representing global confounders (e.g., varying institutional staining protocols, scanner color profiles, and tissue processing artifacts). The causal relationships among the pathological image $X$, the diagnosis $Y$, and the confounder $Z$ are structurally defined by the following topological links:

$X \rightarrow Y$: The true causal effect where genuine morphological features dictate the diagnosis.

$Z \rightarrow X$: The unobserved confounder affects the observed appearance distribution of the WSI, including staining style, scanner-dependent color profiles, tissue-processing artifacts, and other non-diagnostic visual variations. Although these factors do not constitute genuine pathological morphology, they are embedded in the input image $X$. Due to the intrinsic representational bias and limited pathological understanding of neural image encoders, the model may fail to completely separate these $Z$-induced visual variations from true diagnostic evidence.

$Z \rightarrow Y$: Selection bias in data collection (e.g., a specific hospital with a unique staining protocol $Z$ predominantly collects severe malignant cases $Y$), establishing a correlation between the confounder and the diagnostic label.

Based on this causal DAG, the presence of the shared confounder $Z$ inherently derives a spurious link, known in causal graph theory as the backdoor path: $X \leftarrow Z \rightarrow Y$. The presence of this path implies that, in practice, when a model attempts to learn the prediction of $\hat{Y}$ from input $X$, it inadvertently captures correlations confounded by variable $Z$. Consequently, in realistic settings, the resulting observed distribution can be expanded as:

$$\mathrm{P}(\hat{Y} \mid X) = \mathbb{E}_{\mathbb{Z}\sim\mathbb{P}(\mathbb{Z}|\mathbb{X})}\left[P(\hat{Y} \mid X, Z)\right] \quad (2)$$

This equation reveals that the learned $\hat{Y}$ is effectively an approximation biased by $Z$. Driven by the term $P(Z \mid X)$, the model tends to exploit the spurious correlation between $Z$ and $\hat{Y}$, thereby learning a predictive "shortcut" rather than capturing the true causal mechanism defined in Equation (1).

2.2 Micro Confounding and Bias in the MIL Framework

To reveal the formation mechanism of predictive bias, the macro-level process outlined in Section 2.1 is formally decomposed into instance-level variables under the MIL framework.

MIL models are tasked with detecting the presence of positive instances (e.g., localized tumor regions) within a bag containing numerous instances. If at least one positive instance is present, the bag is classified as positive; otherwise, it is classified as negative.

Assuming a WSI is represented as a bag $X = \{x_1, x_2, \dots, x_n\}$ comprising $n$ instance-level patches, with corresponding unobserved instance-level labels $\{y_1, y_2, \dots, y_n\}$ where $y_i \in \{0,1\}$. According to the standard MIL assumption, if at least one positive instance is present, the bag is classified as positive; otherwise, it is classified as negative. The WSI-level bag label $Y \in \{0, 1\}$ can then be strictly formulated as follows:

$$\mathrm{Y} = \begin{cases} 0, \ iff \sum_{i=1}^{n} y_i = 0 \\ 1, \ otherwise \end{cases} \tag{3}$$

In practice, the network cannot directly access the biological ground truth $y_i$. Instead, it approximates the true diagnostic label $Y$ through a parameterized predictive model. Under the standard MIL neural instantiation, the prediction $\hat{Y}$ is derived by utilizing an aggregator $\theta_a(\cdot)$ to fuse instance-level features, followed by a classification module $\theta_c(\cdot)$ to map the aggregated bag-level representation to the final diagnosis:

$$\hat{\mathrm{Y}} = \theta_c\big(\theta_a(x_1, x_2, \dots, x_n)\big) \tag{4}$$

In a perfectly unconfounded scenario, the modules $\theta_a$ and $\theta_c$ would exclusively process genuine pathological morphologies, seamlessly aligning $\hat{Y}$ with the true $Y$.

To mathematically formalize this ideal alignment and the unconfounded causal mechanism, we can expand the true predictive distribution $P(Y|X)$ by marginalizing over the entire latent instance-level label space $\{y_1, y_2, \dots, y_n\}$. According to the law of total probability and the structural characteristics of biological tissues, this marginalization is governed by two critical properties:

1. As established in Equation (3), the WSI-level label $Y$ is strictly and solely determined by the latent instance labels $\{y_i\}$. This renders the raw instance features $\{x_i\}$ conditionally independent of $Y$ given $\{y_i\}$, meaning $P(Y \mid y_1, \dots, y_n, x_1, \dots, x_n) = P(Y \mid y_1, \dots, y_n)$. This term functions mathematically as a deterministic indicator.

2. In real biological tissues, pathological regions exhibit spatial continuity and microenvironmental correlations. The true biological states of the patches are not mutually independent. Therefore, the joint probability $P(y_1, \dots, y_n \mid x_1, \dots, x_n)$ holistically captures the true underlying pathological state driven by the spatially correlated morphological features across the entire WSI.

By incorporating the conditional independence from the deterministic aggregation property directly into the total probability expansion, the ideal, unconfounded MIL predictive distribution can be elegantly formulated as:

$$\begin{aligned} &P(Y|X) \\ &= \sum_{\{y_i\}\in\{0,1\}^n} P(Y \mid y_1, \dots, y_n, x_1, \dots, x_n) P(y_1, \dots, y_n \mid x_1, \dots, x_n) \\ &= \sum_{\{y_i\}\in\{0,1\}^n} P(Y \mid y_1, \dots, y_n) P(y_1, \dots, y_n \mid x_1, \dots, x_n) \end{aligned} \tag{5}$$

This establishes the unconfounded, ideal micro-causal chain: the true instance morphology $x_i$ determines $y_i$, which collectively determines $Y$.

However, the empirically fitted $\hat{Y}$ is inevitably influenced by the macroscopic confounder $Z$. To formalize this microscopic confounding mechanism, the global confounder $Z$ is structurally decomposed into a hierarchical set: $Z = \{Z', z_1, z_2, \dots, z_n\}$.Crucially, these decomposed components strictly correspond to the spurious paths defined in our macroscopic causal DAG:

$\boldsymbol{z_1, z_2, \dots, z_n}$**:** These correspond to the localized manifestations of the $Z \to X$ path at the patch level. Each $z_i$ denotes the non-diagnostic visual variation embedded in the corresponding patch $x_i$, such as local staining fluctuation, scanner-induced color shift, tissue-processing artifact, or background texture bias. Due to the intrinsic representational limitations of neural image encoders, such local confounding factors may be entangled with genuine pathological morphology during feature extraction, leading to biased instance-level representations.

$\boldsymbol{Z'}$**:**This corresponds to the $Z \to Y$ path. $Z'$ represents systemic selection biases injected during the MIL aggregation and classification process (e.g., the model erroneously associating a hospital's specific background stroma color with severe malignancy).

Consequently, from an instantiated perspective, the network's actual prediction process is no longer a pure function of $X$. The feature extraction is coupled with $z_i$, and the classification is modulated by $Z'$. The formulation of the network prediction structurally degenerates into a confounded mapping:

$$\hat{Y} = \theta_c(\theta_a(\{(x_i, z_i)\}_{i=1}^{n}), Z') \tag{7}$$

Building upon the instantiated network architecture, we can now expand the observational predictive distribution $P(\hat{Y}|X)$ for the given input bag $X = \{x_1, x_2, \dots, x_n\}$. By marginalizing over the entire latent confounder space—comprising both the local instance-level artifacts $z_i$ and the global bag-level bias $Z'$—the law of total probability yields:

$$P(\hat{Y} \mid x_1, \dots, x_n) = \sum_{Z'} \sum_{z_1,\dots,z_n} P(\hat{Y} \mid x_1, \dots, x_n, z_1, \dots, z_n, Z') P(z_1, \dots, z_n, Z' \mid x_1, \dots, x_n) \quad (8)$$

Specifically, the term $P(\hat{Y} \mid x_1, \dots, x_n, z_1, \dots, z_n, Z')$ corresponds to the network computation process detailed above, demonstrating that the genuine morphological features $\{x_i\}$ become inevitably entangled with confounding variables during both feature extraction and aggregation. Meanwhile, the term $P(z_1, \dots, z_n, Z' \mid x_1, \dots, x_n)$ reveals that, due to its inherent lack of causal awareness, the model resorts to shortcut learning by directly deriving spurious correlations from the confounders. For instance, the network may infer the hospital-level global staining bias $Z'$ by capturing local color artifacts $z_i$ within a given patch $x_i$, and subsequently exploit this non-causal association to generate the diagnostic prediction $\hat{Y}$.

By comparing the ideal causal predictive distribution in Equation (5) with the instantiated confounded distribution in Equation (8), we can formally define the core optimization objective of a causal MIL framework. The fundamental goal is to eliminate the predictive bias by minimizing the discrepancy between the network's confounded empirical prediction $P(\hat{Y}|X)$ and the true, unconfounded causal distribution $P(Y|X)$:

$$\min\left|P(\hat{Y}|X) - P(Y|X)\right| = min\left|\sum_{Z',\{z_i\}} P(\hat{Y}|\{x_i\},\{z_i\},Z')P(\{z_i\},Z'|\{x_i\}) - \sum_{\{y_i\}} P(Y|\{y_i\})P(\{y_i\}|\{x_i\})\right| \quad (9)$$

This mathematical expansion transparently exposes the structural root of the predictive bias. The right-hand term represents the ideal causal mechanism defined in Equation (5), which is driven exclusively by the latent instance-level diagnostic states $\mathbf{y} = \{y_i\}_{i=1}^{N}$. In contrast, the left-hand term represents the network's empirically fitted observational prediction, where the prediction is additionally modulated by the posterior distribution of the latent confounders $\mathbf{Z} = (\{z_i\}_{i=1}^{N}, Z')$. In the SCM, the predictive bias arises because the latent confounders are structurally associated with both the observed WSI features and the diagnostic outcome, corresponding to the open backdoor path $\mathbf{X} \leftarrow \mathbf{Z} \rightarrow Y$.

2.3 The basic process of front door intervention.

Therefore, this paper introduces causal intervention to reduce the confounding bias revealed in Equation (9). In Pearl's causal framework, the conventional conditional probability $P(\hat{Y} \mid X)$ describes a passive observation of $X$, whose prediction tends to absorb spurious correlations induced by the unobserved confounder $Z$. In contrast, $P(\hat{Y} \mid do(X))$ describes an active intervention on $X$, where the natural image-generation mechanism is replaced by a controlled feature-generation mechanism.

$do(X)$ is introduced to optimize $X$ into a purified feature distribution. This purified distribution is expected to be decoupled from the visual appearance of the original WSI and the acquisition-related confounding factors, while preserving diagnostically meaningful morphological features. In the SCM, this operation aims to cut off all incoming edges to $X$, namely blocking the spurious backdoor path $Z \rightarrow X$, thereby making $X$ strictly independent of the unobserved confounder $Z$. The mediator variable $M$ is introduced as the concrete realization of $do(X)$, where $M$ denotes the purified feature representation generated after intervening on $X$.

Since $Z$ cannot be directly measured in practical WSI diagnosis, directly estimating $P(\hat{Y} \mid do(X))$ through back-door adjustment is infeasible. In the presence of unobserved confounders, front-door intervention provides a more feasible solution. As shown in Figure 1.c, this method identifies the causal effect through an observable mediator variable $M$, which transmits diagnostic information from $X$ to $Y$. Under the deterministic mapping assumption of neural-network-based feature extraction, the standard front-door adjustment formula can be simplified as an expectation over the global input prior distribution:

$$P(\hat{Y}|do(X)) = \mathbb{E}_{\mathbb{X}' \sim \mathbb{P}(\mathbb{X})}[P(\hat{Y} \mid M(X), X')] \quad (10)$$

2.4 Mechanism of Reducing Prediction Bias in Dual-Pathway Model

After obtaining the computable front-door intervention distribution in Eq. (10), we further abstract a common computational form shared by front-door intervention-based MIL methods. We argue that the practical essence of the front-door intervention operator in MIL can be reduced to a dual-channel processing mechanism. Specifically, the same WSI-level input $X$ is processed from two different perspectives:

$$X_1 = \Phi_1(X), \qquad X_2 = \Phi_2(X), \quad (11)$$

where $\Phi_1(\cdot)$ and $\Phi_2(\cdot)$ denote two channel-specific transformation functions. Here, $X_1$ and $X_2$ are the channel-level representations of the same WSI bag $X$. The two channels are not required to be individually causal; they may both be observational feature channels. The subsequent fusion and disentanglement of these two representations can induce the causal effect of $do(X)$, yielding $P(\hat{Y} \mid do(X))$.

Thus, the mediator variable in the front-door formula can be written as:

$$M = \Psi(X_1, X_2), \quad (12)$$

where $\Psi(\cdot,\cdot)$ denotes a learnable operator for constructing the interventional mediator representation. Under the ideal information-preserving scenario, $M$ retains the diagnosis-relevant information contained in the joint representation $(X_1, X_2)$.

The information obtained by the two channels under different processing mechanisms should provide different diagnostic perspectives for the same WSI input, emphasizing different aspects of pathological morphology, such as spatial organization, texture patterns, frequency-domain variations, or contextual dependencies. The information contained in these different perspectives should also be non-overlapping diagnostic information.

In this paper, conditional Shannon entropy $H(\cdot|\cdot)$ is introduced to formalize this point. To illustrate why such a dual-channel structure can outperform a single-channel MIL model, we take $X_1$ as a representative single channel in the following discussion.

Specifically, $H(Y \mid X_1)$ measures the remaining uncertainty of the ideal WSI-level diagnostic outcome $Y$ after observing the first-channel representation $X_1$, while $H(Y \mid X_1, X_2)$ measures the remaining diagnostic uncertainty after jointly observing both channel representations. The diagnosis-relevant non-redundancy condition is formulated as

$$H(Y \mid X_1) - H(Y \mid X_1, X_2) > 0. \quad (13)$$

Equation (13) states that, after $X_1$ is known, the addition of $X_2$ reduces the diagnostic uncertainty, i.e., the second channel still provides additional information about the ideal WSI-level diagnostic outcome $Y$.

To demonstrate that the causal intervention is effective, we aim to show that the theoretically optimal front-door interventional discrepancy is smaller than the theoretically optimal observational discrepancy.The KL divergence is introduced as a distributional discrepancy measure $D_{KL}(\cdot \,||\, \cdot)$.

$$\begin{aligned} &\left|P\left(\hat{Y}\middle|do(X)\right) - P(Y|X)\right| - \left|P(\hat{Y}|X) - P(Y|X)\right| \\ &= \mathbb{E}_{X' \sim P(X)}[D_{\mathrm{KL}}(P(Y \mid X') \parallel P(\hat{Y} \mid do(X')))] \\ &\qquad - \mathbb{E}_{X' \sim P(X)}\left[D_{\mathrm{KL}}\left(P(Y \mid X') \parallel P(\hat{Y} \mid X')\right)\right] \\ &= H(Y|X_1, X_2) - H(Y|X_1) < 0 \end{aligned} \quad (14)$$

Equation (14) shows that, under the diagnostic complementarity condition and the ideal information-preserving mediator assumption, the KL divergence between the theoretically optimal dual-channel interventional prediction and the ideal unconfounded causal predictive distribution is smaller than that between the theoretically optimal single-channel observational prediction and the ideal distribution. That is, the interventional prediction $P(\hat{Y} \mid do(X))$ is expected to be closer to the ideal unconfounded causal predictive distribution $P(Y \mid X)$ than the original observational prediction $P(\hat{Y} \mid X)$. This agrees with the optimization direction of causal MIL, namely reducing the discrepancy between the model prediction and the ideal causal target through feature-space intervention. Therefore, under the stated assumptions, the dual-channel front-door intervention provides a computable mechanism for mitigating confounding-induced prediction bias.

### 2.5 Additional conditional assumptions under non ideal conditions

The analysis in Section 2.4 describes an ideal setting in which the dual-channel representation contains diagnosis-relevant non-redundant information and the mediator $\boldsymbol{M} = \boldsymbol{\Psi}(\boldsymbol{X_1}, \boldsymbol{X_2})$ preserves such information during fusion. However, in practical MIL models, these two requirements are not automatically satisfied. Therefore, we propose that abstracting the front-door intervention method into a dual-channel representation entails two additional conditions.

First, each channel should possess independent and sufficient diagnostic capability. This condition requires that neither channel degenerates into purely noisy or non-diagnostic transformations. The remaining diagnostic uncertainty of a single channel, $\boldsymbol{H}(\boldsymbol{Y} \mid \boldsymbol{X_k})$, should be sufficiently small, such that each channel, when used alone, can support baseline-level WSI classification. This requirement ensures that both $\boldsymbol{X_1}$ and $\boldsymbol{X_2}$ retain sufficient pathological evidence to predict the ideal WSI-level diagnostic outcome $\boldsymbol{Y}$.

Second, under the premise of independent diagnostic capability, a larger inter-channel feature discrepancy can serve as an indicator of stronger potential diagnostic complementarity. A larger discrepancy between the two channels indicates that they process the same WSI input from more distinct feature perspectives. For example, image-domain and frequency-domain transformations are unlikely to form strict subset relations, because they emphasize different aspects of pathological morphology. When both channels possess independent and sufficient diagnostic capability, non-diagnostic information is less likely to dominate either channel—otherwise, excessive non-diagnostic information would impair the single-channel diagnostic performance. Therefore, a larger inter-channel discrepancy is more likely to correspond to non-overlapping diagnostic information, and the diagnostic complementarity between the two channels is positively correlated with $\boldsymbol{H}(\boldsymbol{X_2} \mid \boldsymbol{X_1})$.

**2.6 Network Instantiation of the Additional Assumptions**

Given a WSI bag $X = \{x_i\}_{i=1}^{N}$, two feature pathways extract the patch-level representations

$$h_{1i} = \Phi_1(x_i), h_{2i} = \Phi_2(x_i) \tag{15}$$

Each pathway follows the general MIL classification structure. For the $k$-th pathway, $k \in \{1,2\}$, the attention score and attention weight of the $i$-th patch are calculated as

$$e_{ki} = \frac{q_k^\top h_{ki}}{\sqrt{d}}, a_{ki} = \frac{\exp(e_{ki})}{\sum_{j=1}^{N} \exp(e_{kj})} \tag{16}$$

where $q_k$ is a learnable attention vector and $d$ is the feature dimension. The corresponding WSI-level representation is

$$z_k = \sum_{i=1}^{N} a_{ki} h_{ki} \tag{17}$$

Thus, both pathways contain patch-level feature extraction, attention-based aggregation, and WSI-level representation learning, and each can independently support slide-level classification when it retains sufficient pathological evidence. This satisfies the first assumption that both pathways possess independent diagnostic capability.

The two pathways are then fused through a difference-driven attention mechanism. Let

$$\bar{h}_{ki} = \frac{h_{ki}}{\| h_{ki} \|_2} \tag{18}$$

denote the normalized patch representation, and let the normalized first-pathway WSI representation be expressed as

$$\bar{z}_1 = \sum_{j=1}^{N} a_{1j} \bar{h}_{1j} \tag{19}$$

For each second-pathway patch feature, its discrepancy from the first-pathway WSI-level representation is defined via the inner product:

$$d_i = 1 - \bar{h}_{2i}^\top \bar{z}_1 \tag{20}$$

A larger $d_i$ indicates that the second-pathway patch feature is more different from the dominant WSI-level representation captured by the first pathway. Since $\bar{z}_1$ is an attention-weighted combination of the first-pathway patch features, this discrepancy can be expanded as

$$d_i = 1 - \bar{h}_{2i}^\top \sum_{j=1}^{N} a_{1j} \bar{h}_{1j} = \sum_{j=1}^{N} a_{1j} \left(1 - \bar{h}_{2i}^\top \bar{h}_{1j}\right) \tag{21}$$

Thus, the discrepancy between a second-pathway patch feature and the first-pathway WSI-level representation is exactly the attention-weighted average discrepancy between that feature and the first-pathway patch-level representations. Hence, a larger distance from $\bar{h}_{2i}$ to $\bar{z}_1$ indicates a larger average feature difference from the diagnostically emphasized patch distribution of the first pathway. The discrepancy scores are converted into fusion weights through softmax normalization:

$$\alpha_i = \frac{\exp(d_i)}{\sum_{j=1}^{N} \exp(d_j)}, \tag{22}$$

so that second-pathway features that differ more from the first-pathway representation receive larger fusion weights. The attention-weighted second-pathway representation is

$$r_2 = \sum_{i=1}^{N} \alpha_i h_{2i}, \tag{23}$$

and the interventional mediator is constructed by fusing this representation with the first-pathway WSI-level feature:

$$M = z_1 + W_f r_2, \tag{24}$$

where $W_f$ is a learnable feature transformation.

This thus satisfies the second assumption: when both pathways independently preserve diagnostic information, a larger feature difference indicates that the second pathway is more likely to provide pathological evidence not already emphasized by the first pathway. The difference-driven attention mechanism assigns greater contribution to these non-redundant features and incorporates them into the mediator $M$.